# Contextual Bandits for adapting to changing User preferences over time

Dattaraj Jagdish Rao (dattaraj_rao@persistent.com)
Persistent Systems, Goa INDIA

**ABSTRACT**

Traditional Machine Learning (ML) systems work well when there is a classification or regression problem with static datasets. However, in the real world we don't have the luxury of static datasets as the environment from which data is collected changes continuously. Contextual bandits provide an effective way to model the dynamic data problem in ML by leveraging online (incremental) learning to continuously adjust the predictions based on changing environment. We explore details on contextual bandits, an extension to the traditional reinforcement learning (RL) problem and build a novel algorithm to solve this problem using an array of action-based learners. We apply this approach to model an article recommendation system using an array of stochastic gradient descent (SGD) learners to make predictions on rewards based on actions taken. We then extend the approach to a publicly available MovieLens dataset and explore the findings.

The contributions of this work are four-fold.

- First, we make available a simplified simulated dataset showing varying user preferences over time and how this can be evaluated with static and dynamic learning algorithms. This dataset made available as part of this research is intentionally simulated with limited number of features and can be used to evaluate different problem-solving strategies.

- We will build a classifier using static dataset and evaluate its performance on this dataset. We show limitations of static learner due to fixed context at a point of time and how changing that context brings down the accuracy.

- Next we develop a novel algorithm for solving the contextual bandit problem. Similar to the linear bandits [1], this algorithm maps the reward as a function of context vector but uses an array of learners to capture variation between actions/arms. We develop a bandit algorithm using an array of stochastic gradient descent (SGD) learners – with separate learner per arm.

- Finally, we will apply this contextual bandit algorithm to predicting movie ratings over time by different users from the standard Movie Lens dataset and demonstrate the results.

The author makes the algorithm and dataset from this paper available open source on GitHub. The code is packaged as a reusable Python class following the Object-Oriented Programming (OOP) paradigm. The code is made open source and can easily be adopted to a contextual bandit problem in a different domain.

**CCS CONCEPTS**
• Information systems ~ Information retrieval ~ Retrieval tasks and goals ~ Recommender systems
• Computing methodologies ~ Machine learning ~ Machine learning algorithms

**KEYWORDS**
contextual bandits, recommender systems, reinforcement learning, online learning algorithm, rating prediction

# 1  Introduction: Reinforcement Learning to Contextual bandits

Traditional Machine Learning (ML) systems work well when there is a classification or regression problem with static dataset. However, in the real world we don't have the luxury of static datasets as the environment from which data is collected keeps changing. For example, if we model an online article recommendation system to capture user preferences, over time the user preferences may change, and the model may not be valid in a few months. Many environmental changes like weather, social factors or major events affect user preferences and it's almost impossible to accurately capture these in a model. For example, if we take case of user's readings online articles, probably the general tendency is to opt for movie or sports related articles – at least for some age groups. However, due to the recent unfortunate pandemic of COVID-19, we will see a big spike in users viewing health and medical related articles. To capture these changing preferences, we will have to continuously retrain our model and redeploy into production. Alternate approach is to have an adaptive learning algorithm that continuously learns and adjusts itself to changing preferences. A special category of ML algorithms that handle this scenario is known as reinforcement learning (RL). In RL, we develop an agent to learn by "observing" an environment rather than from a static dataset. RL is considered to be more of a true form of Artificial Intelligence (AI) – because its analogous to how we, as humans, learn new things – observing and learning by trial and error. Although, RL has been studied by researchers for decades now, a particular formulation of RL that is fast gaining popularity is called Multi-armed bandits (MAB). In this article, we will explore Multi-armed Bandits and understand how these can be applied to areas like improving design of websites and making clinical trials more effective.

As shown in figure 1 below, traditional RL problem is modelled as an environment with a state (S1) which is observed by our agent and changed to state (S2) by taking an action (A). The action transitions state of the environment from (S1) to (S2) and in return the agent gets reward (R). The reward may be positive or negative. Over a series of such trials and errors the agent learns an optimal policy to take actions based on state which maximize the long-term rewards. Example of this could be a game of chess where actions taken by the player change the state of the board and there may be immediate rewards like killing or losing a piece and long-term reward of winning or losing the game. RL is heavily used in gaming industry and you can imagine these agents and environments becoming more and more complex.

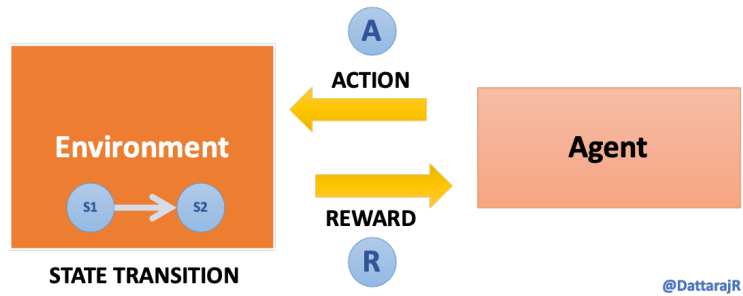

**Figure 1: Pure Reinforcement Learning**

A simpler abstraction of the RL problem is the **multi-armed bandit** problem. A multi-armed bandit problem does not account for the environment and its state changes. As shown in figure 2 below, here the agent only observes the actions it takes and rewards it receives and tries to devise the optimal strategy. The idea in solving multi-armed bandit problems is to try and explore the action space and understand the distribution of the unknown rewards function. The name "**bandit**" comes from the analogy of Casinos where we have multiple slot machines and we have to decide if we should continue playing a single machine (exploitation) or moving to a new machine (exploration) – all with objective of maximizing our winnings. We don't know anything about the state of these machines – only the actions we take and rewards we get. Here we need to decide between multiple choices purely by taking actions and observing the returns. These algorithms ultimately try and do a trade-off between **exploration** and **exploitation** to identify optimal strategy.

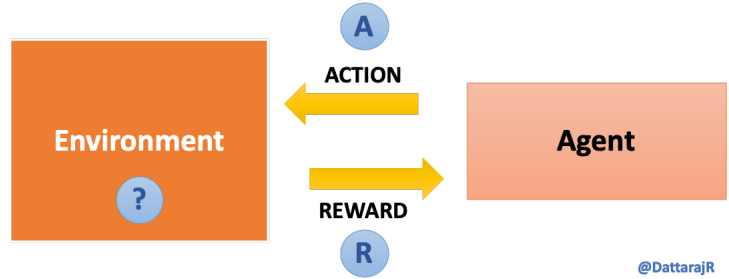

**Figure 2: Multi-armed Bandit problem**

A drawback of multi-armed bandits is that they totally ignore the state of the environment. The state can provide very valuable insights that can help learn a policy much faster and more effectively. Incorporating some element of state from the problem environment has given rise to a new set of algorithms **contextual bandits,** shown in figure 4 below. Here, instead of randomly managing the exploration vs exploitation trade-off, we try and obtain some **context** about the environment and use that to manage the action. Context here is different than the State we talked about for a traditional RL problem earlier. Context is just some knowledge that we have of the environment that helps us take an action. For example, if the website B uses design elements that are known to be preferred by people under the age of 30, then for users under 30 we may choose to "exploit" this Context information and always show the website B.

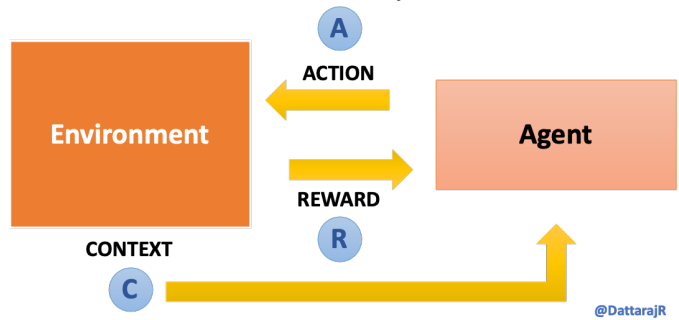

**Figure 3: Contextual Bandits**

Contextual Bandits [1][2][3][4][5][6] are extensively used in areas like personalized advertisement to show relevant ads to consumers. We can see that in areas like website design and clinical trials also, having the right Context can go a long way in making our actions more effective. In the website design example, Context is typically modelled as a vector that contains information like user location, browser history, browser type, etc. For clinical trials we can have even more granular information like patient age, sex, medical history, etc. For clinical trials, there may even be some pre-checks and screenings used to generate Context information that will help in future to make an informed decision.

Contextual bandits are also being used in **recommender systems** [8][9] where we predict items to recommend to a particular user. Examples are product recommendations on Amazon or movie recommendations on Netflix. Traditionally, companies have been using techniques like content and collaborative filtering for providing recommendation. **Content filtering** tries to find items similar to the one's user has liked before using features genre, actors, country, release year, etc. **Collaborative filtering** on the other hand uses 'wisdom of crowd' approach by finding users similar to current user and find items they liked which current user will mostly like. Content filtering needs lot of feature engineering and collaborative filtering has a cold start problem. Particularly when we have new users or new movies, it's difficult to get good recommendations. More recently, companies have started exploring contextual and multi-armed bandits for improving their recommendations. Having user context is an invaluable information that can help to provide more relevant recommendations. This is particularly important when there is a constantly changing user database and items catalogue – as in Netflix or Amazon case.

## 2  Problem and the Dataset

We will look at an article recommendation problem where we get 2 features of the user as context and recommend an article in one of 4 categories – news, movies, sports or health. Based on recommendation and the state of mind of the user he or she may choose to accept the article by clicking on it or ignore it and surf away. Each of these scenarios is captured by a reward value of 1 for click or 0 for otherwise. This is a highly simplified and simulated dataset, but we will use it to demonstrate how user preferences change over time and evaluate multiple algorithms. Our algorithm and implementation will be generic enough to scale to much bigger datasets. Figure 4 below shows how the data looks like. We have 5000 rows in our dataset.

| Gender | Age | Recommendation | Reward |
|---|---|---|---|
| m | 29 | health | 0 |
| m | 28 | movies | 1 |
| m | 34 | news | 1 |
| m | 36 | health | 0 |
| f | 24 | news | 0 |
| f | 61 | sports | 0 |
| f | 30 | sports | 1 |
| f | 18 | news | 0 |
| m | 79 | movies | 0 |
| m | 13 | sports | 1 |
| m | 85 | news | 0 |
| m | 83 | health | 1 |
| f | 19 | news | 0 |
| m | 87 | health | 1 |
| m | 30 | health | 0 |
| m | 56 | movies | 0 |
| m | 73 | news | 0 |
| f | 59 | movies | 1 |
| m | 28 | sports | 1 |
| m | 71 | health | 1 |
| f | 60 | news | 0 |

**Figure 4: Snapshot of the dataset**

Mapping this dataset to our contextual bandits problem in figure 3 – we have vector of Age and Gender as context and accordingly we take an action (pull an arm) to select an article. Based on this we get a reward which is binary and signifies 1 for click and 0 if the article was ignored.

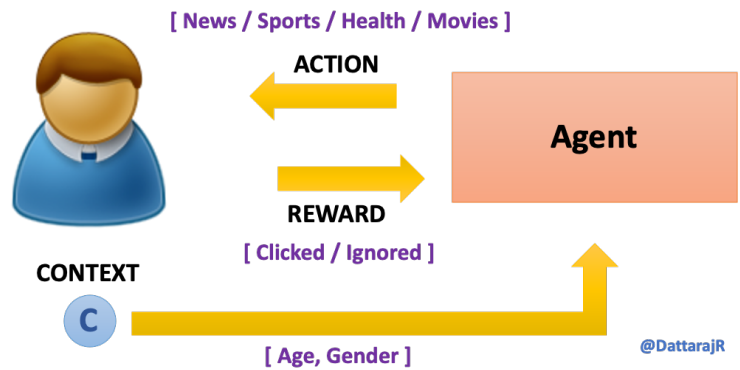

**Figure 5: Contextual bandits problem setting**

## 3  Solving with a static Classifier

Let's use a static classifier to train on the data and use it to predict the reward. Since the reward can either be 0 or 1 we will use a simple binary classifier. We will take the Age, Gender and the recommendation (arm pulled) as input and predict the reward. as Our dataset has 5000 rows. We will use first 500 to train a decision tree classifier and predict on the remaining 4500 points. Since this is a sequential decision-making problem, we will consider the model accuracy gradually over time. For each set of 20 data points, we will predict the reward and compare with the actual reward obtained. We will plot this accuracy over time to see how well our classifier is doing. Below figure 6 shows this accuracy plot for decision tree classifier trained on 500 points. Since we are grouping 20 points at a time to calculate moving accuracy, we see a total of 250 accuracy values on chart for total 5000 points. Green points indicate accuracy values above 70% and red are lesser.

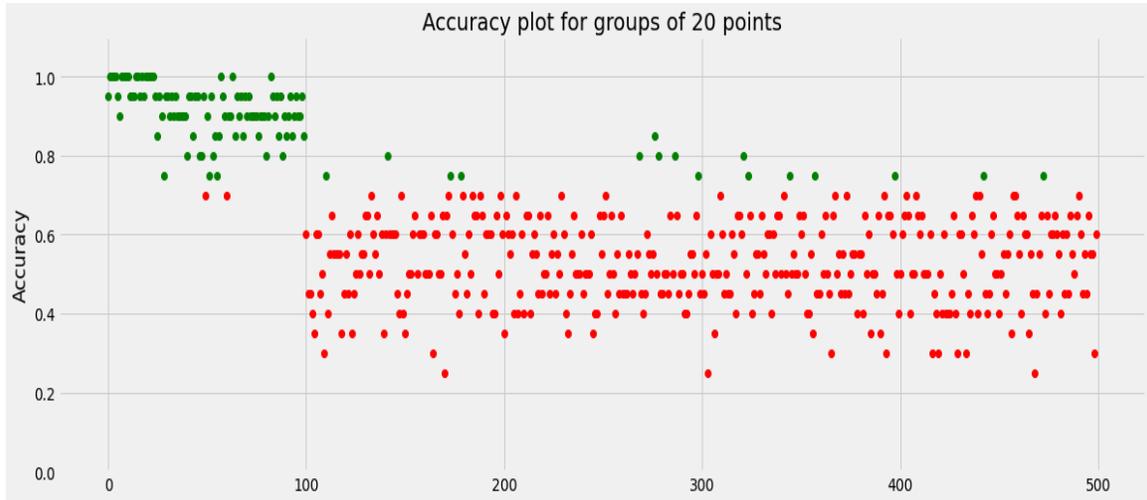

**Figure 6: Using a static learning Classifier – Average accuracy = 70.1%**

From figure 6 we see that our predictor works well for initial set of points and then accuracy falls. This is typically when some major events influence user behavior so that the preferences change. Our static classifier is not able to adapt to these changes in time. This is where incremental or online learning will come into play. If we can learn continuously from the data, we should be able to adapt to the changing user preferences.

## 4   Solving with a single Online Learning model

Let's use a single predictor model that learns continuously on new data. We will apply the stochastic gradient descent (SGD) learner and fit it partially on a set of 20 data points each time and calculate accuracy for next 20 points. Below is chart for a single online learning SGD model that continuously learns on new data. Average accuracy for this learner comes to 66.8%. Although we are using online learning based on context and keep continuously adapting the reward predictor, we do not see a big accuracy number. This leads to the thinking that maybe each action or arm needs a dedicated separate predictor model. This is how we evolve the new algorithm.

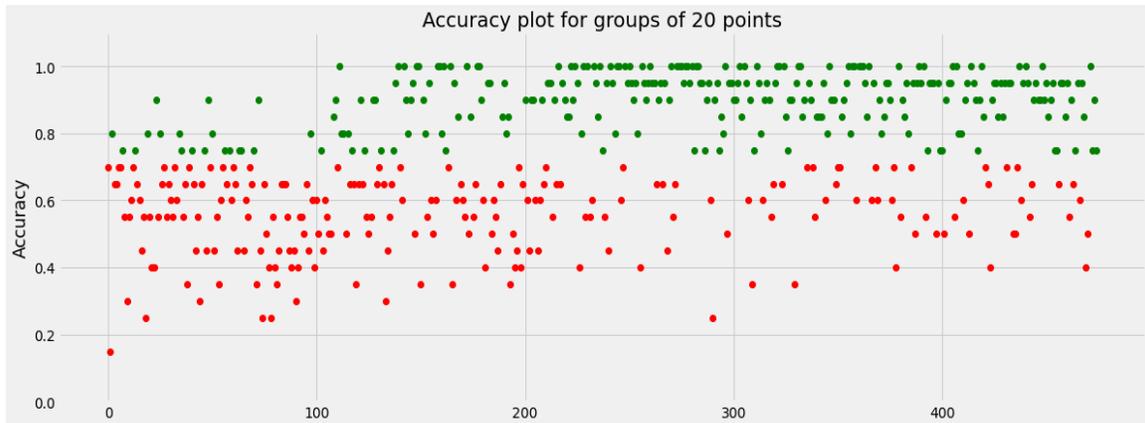
**Figure 7: Using a single online learning SGD model – Average accuracy = 74.9%**

We see that despite using an online learning classifier we do not see much improvement in accuracy. Interestingly in a static classifier we saw that initial data points showed more accuracy while later points showed decline in accuracy. On other hand we see that a single online learning classifier shows better results later on.

## 5   Building an array of bandit learners

We developed an algorithm inspired by linear bandits [1] where we would model the sequential decision-making problem with contextual bandits. The idea is to train a separate SGD classifier per recommendation or per arm of the bandit. We will try and express our action (arm) as a function of the context parameters.

x – Each data point from 1 to N (5000 in our case)
$C_x$ – Context vector for each datapoint
$A_x$ – Action taken, or arm pulled for each datapoint
$R_x$ – Reward obtained by taking $A_x$
Expected reward from arm = E [ Rx | $C_x$ , $A_x$ ] = function (Context $C_x$ , Arm pulled $A_x$)
Arm Learners = Array of SGD classifiers for arms
$SGD_{arm}$ = SGD classifier for $A_x$

Below is a high-level overview of the algorithm.

**ALGORITHM 1**: Array of bandit learners

**define array of bandit learners – dictionary {A}**

**for each record in dataset:**

    **get the context vector $C_x$**

    **normalize $C_x$ so values are between 0 and 1**

    **get the action $A_x$**

    **get the reward $R_x$**

    **normalize $A_x$ and $R_x$ if needed**

    **if learner $SGD_{arm}$ does not exist in dictionary {A}**

        **create $SGD_{arm}$ learner for arm $A_x$**

        **dictionary {arm} = $SGD_{arm}$**

    X = ($C_x$, $A_x$) and Y = $R_x$
    partial fit $SGD_{arm}$ for X, Y
    add $SGD_{arm}$ to array of arms
    go to next record
   end if
   get prediction $P_x$ using $SGD_{arm}$
   compare $P_x$ and $R_x$
   X = ($C_x$, $A_x$) and Y = $R_x$
   partial fit $SGD_{arm}$ for X, Y
end for

Implementing the above algorithm for our dataset we get results shown in figure 8 below. We see that the average accuracy is higher at 85.5%. The key differentiators here are to have separate learner for each bandit arm and keep adjusting this learner with new values as they come in. After observing the data from each arm, the model starts to generalize well and make accurate predictions. The model will have a cold start problem and will need enough data to train each arm before the arm learner starts making predictions. The array of models can accurately map to the data and adjust for changing patterns in user preferences.

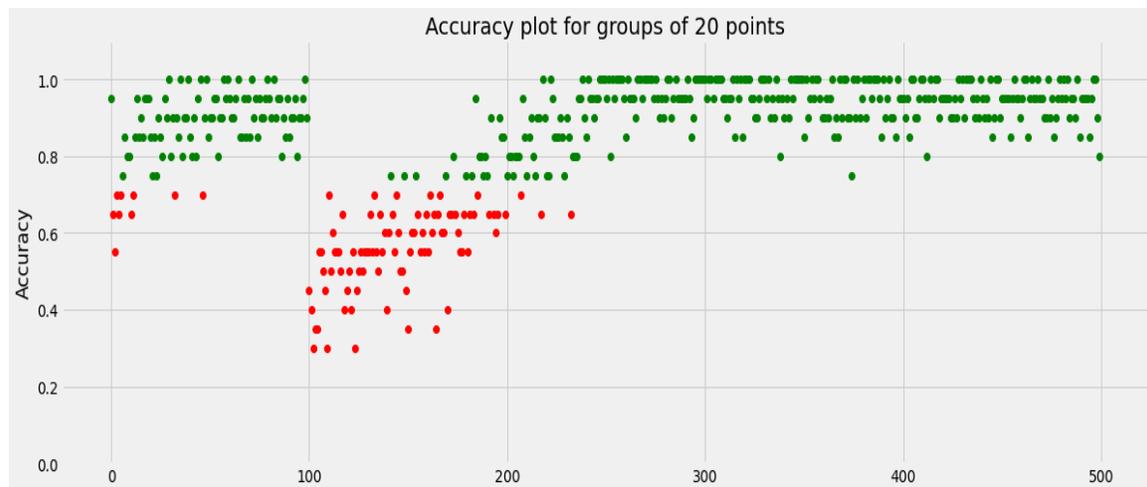

**Figure 8: Using array of bandit SGD learners – Average accuracy = 85.5%**

## 6  Applying the algorithm for recommendations – Movie Lens

We will now apply the contextual bandits model to predict rankings that will be given by users to movies. Traditionally collaborative filtering [11] methods are popular for this use case, but we will frame this as a bandits problem. The analysis is done of the popular movie review dataset called MovieLens [10]. MovieLens data sets were collected by the GroupLens Research Project at the University of Minnesota. The data was collected through the MovieLens web site (movielens.umn.edu) during the seven-month period from September 19th, 1997 through April 22nd, 1998. This data has been cleaned up – users who had less than 20 ratings or did not have complete demographic information were removed from this data set. This data set consists of 100,000 ratings (1-5) from 943 users on 1682 movies, where each user has rated at least 20 movies and simple demographic info for the users (age, gender, occupation, zip).

Data from MovieLens is distributed in multiple files connected on specific columns as in a relational database. We aggregated data from following data files: user (u.user), movies (u.item) and ratings (u.data) on fields 'movie_id' and 'user_id'. We also do some feature engineering to change fields like 'sex', 'occupation' and 'zip_code' into categorical fields for better analysis. We will also remove any fields with null values for ratings. The code for this feature engineering is available on GitHub [5]. Figure 9 below has snapshot of the cleaned data.

| movie_id | title | user_id | rating | unix_timestamp | age | sex | occupation | zip_code | Action | Adventure | Animation | Children | Comedy |
|---|---|---|---|---|---|---|---|---|---|---|---|---|---|
| 255 | My Best Friend's Wedding (1997) | 259 | 4 | 874724710 | 21 | 1 | 18 | 367 | 0 | 0 | 0 | 0 | 1 |
| 286 | English Patient, The (1996) | 259 | 4 | 874724727 | 21 | 1 | 18 | 367 | 0 | 0 | 0 | 0 | 0 |
| 298 | Face/Off (1997) | 259 | 4 | 874724754 | 21 | 1 | 18 | 367 | 1 | 0 | 0 | 0 | 0 |
| 185 | Psycho (1960) | 259 | 4 | 874724781 | 21 | 1 | 18 | 367 | 0 | 0 | 0 | 0 | 0 |
| 173 | Princess Bride, The (1987) | 259 | 4 | 874724843 | 21 | 1 | 18 | 367 | 1 | 1 | 0 | 0 | 1 |
| 772 | Kids (1995) | 259 | 4 | 874724882 | 21 | 1 | 18 | 367 | 0 | 0 | 0 | 0 | 0 |
| 108 | Kids in the Hall: Brain Candy (1996) | 259 | 4 | 874724882 | 21 | 1 | 18 | 367 | 0 | 0 | 0 | 0 | 1 |
| 288 | Scream (1996) | 259 | 3 | 874724905 | 21 | 1 | 18 | 367 | 0 | 0 | 0 | 0 | 0 |
| 928 | Craft, The (1996) | 259 | 4 | 874724937 | 21 | 1 | 18 | 367 | 0 | 0 | 0 | 0 | 0 |
| 117 | Rock, The (1996) | 259 | 4 | 874724988 | 21 | 1 | 18 | 367 | 1 | 1 | 0 | 0 | 0 |
| 200 | Shining, The (1980) | 259 | 4 | 874725081 | 21 | 1 | 18 | 367 | 0 | 0 | 0 | 0 | 0 |
| 405 | Mission: Impossible (1996) | 259 | 3 | 874725120 | 21 | 1 | 18 | 367 | 1 | 1 | 0 | 0 | 0 |
| 1074 | Reality Bites (1994) | 259 | 3 | 874725264 | 21 | 1 | 18 | 367 | 0 | 0 | 0 | 0 | 1 |
| 176 | Aliens (1986) | 259 | 4 | 874725386 | 21 | 1 | 18 | 367 | 1 | 0 | 0 | 0 | 0 |
| 357 | e Flew Over the Cuckoo's Nest (197 | 259 | 5 | 874725485 | 21 | 1 | 18 | 367 | 0 | 0 | 0 | 0 | 0 |
| 210 | iana Jones and the Last Crusade (19 | 259 | 4 | 874725485 | 21 | 1 | 18 | 367 | 1 | 1 | 0 | 0 | 0 |
| 687 | McHale's Navy (1997) | 851 | 2 | 874728168 | 18 | 1 | 13 | 250 | 0 | 0 | 0 | 0 | 1 |
| 284 | Tin Cup (1996) | 851 | 3 | 874728338 | 18 | 1 | 13 | 250 | 0 | 0 | 0 | 0 | 1 |
| 696 | City Hall (1996) | 851 | 3 | 874728338 | 18 | 1 | 13 | 250 | 0 | 0 | 0 | 0 | 0 |
| 295 | Breakdown (1997) | 851 | 5 | 874728370 | 18 | 1 | 13 | 250 | 1 | 0 | 0 | 0 | 0 |
| 473 | James and the Giant Peach (1996) | 851 | 4 | 874728396 | 18 | 1 | 13 | 250 | 0 | 0 | 1 | 1 | 0 |
| 544 | to Do in Denver when You're Dead | 851 | 4 | 874728396 | 18 | 1 | 13 | 250 | 0 | 0 | 0 | 0 | 0 |
| 290 | Fierce Creatures (1997) | 851 | 4 | 874728430 | 18 | 1 | 13 | 250 | 0 | 0 | 0 | 0 | 1 |
| 147 | Long Kiss Goodnight, The (1996) | 851 | 4 | 874728461 | 18 | 1 | 13 | 250 | 1 | 0 | 0 | 0 | 0 |
| 121 | Independence Day (ID4) (1996) | 851 | 4 | 874728565 | 18 | 1 | 13 | 250 | 1 | 0 | 0 | 0 | 0 |
| 717 | Juror, The (1996) | 851 | 3 | 874728598 | 18 | 1 | 13 | 250 | 0 | 0 | 0 | 0 | 0 |
| 1040 | Two if by Sea (1996) | 712 | 4 | 874729682 | 22 | 0 | 18 | 410 | 0 | 0 | 0 | 0 | 1 |
| 220 | Mirror Has Two Faces, The (1996) | 712 | 5 | 874729682 | 22 | 0 | 18 | 410 | 0 | 0 | 0 | 0 | 1 |
| 510 | Magnificent Seven, The (1954) | 712 | 2 | 874729749 | 22 | 0 | 18 | 410 | 1 | 0 | 0 | 0 | 0 |
| 50 | Star Wars (1977) | 712 | 4 | 874729750 | 22 | 0 | 18 | 410 | 1 | 1 | 0 | 0 | 0 |
| 417 | Parent Trap, The (1961) | 712 | 4 | 874729750 | 22 | 0 | 18 | 410 | 0 | 0 | 0 | 1 | 0 |
| 731 | Corrina, Corrina (1994) | 712 | 5 | 874729750 | 22 | 0 | 18 | 410 | 0 | 0 | 0 | 0 | 1 |
| 623 | Angels in the Outfield (1994) | 712 | 4 | 874729778 | 22 | 0 | 18 | 410 | 0 | 0 | 0 | 1 | 1 |
| 385 | True Lies (1994) | 712 | 5 | 874729778 | 22 | 0 | 18 | 410 | 1 | 1 | 0 | 0 | 1 |

**Figure 9: Snapshot of the MovieLens dataset**

We see that the dataset has distinct genres for each movie and each movie may be part of multiple genres. Instead of taking each individual movie name as a bandit arm – we will build a movie rating predictor based on genre. So, in our setup the user features – age, sex, zip_code will be context vector. For each movie, we will get the context vector and normalize it. Normalization usually involves converting the value to a number between 0 to 1 so that computations become faster. Then we will pass this context vector to our bandit learner class along with the arm to make reward prediction for – that is representing action to be taken. Our bandit algorithm will predict the movie rating which we will compare with the actual movie rating. Idea is to learn the probability for the rating based on the given context. Since our action / recommendation / bandit arm is the genre, we will share the rating for a movie with every genre that movie is categorized under. This is the most important assumption and limitation of this approach. If a movie – say 'Finding Nemo' is categorized as Children's and Animation – we will share the rating for each user to each of the genres. This way we get new data points for every movie for each genre it falls under. If we run the test for 100 movies, we will get training records in multiples of 100 due to the many to many relationships between genre and movies. We will use the first 500 records exclusively for training and then take the next 2500 records and make predictions. We will continue to train on the real value of rating after storing our prediction. We get an array of actual ratings per movie per genre and an array of predicted ratings. We will compare the 2 and create a matrix of accuracy. We will calculate accuracy over time using a sliding window on actual and prediction arrays and plot this accuracy on a chart shown in figure 10 below. We see a pretty good accuracy of 80% on the dataset. Again, keep in mind that we are actually continuously learning, and our learner keeps adapting to changes in movie ratings over time, This helps it to capture key changes in preferences due to social and environmental factors.

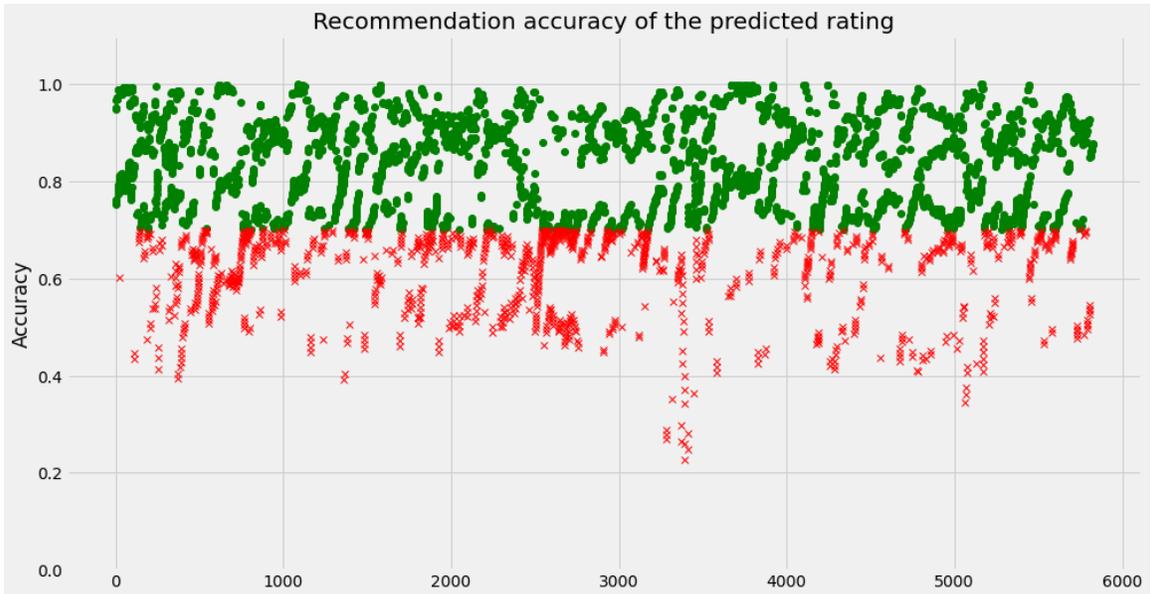
**Figure 10: Predicting movie ratings from observed data – Average Accuracy = 82%**

# 7 Comparing MovieLens results from static regression models

We will now run some static learning models on the MovieLens dataset and compare results with our bandit learner algorithm. We took first 1000 records as our training set and trained a linear regression and decision tree classifier. We used the rest of data as test set and evaluated our predicted ranking. Since ranking is a number between 1 and 5, we used root mean square error (RMSE) as the accuracy measure – lower the error better is the learning. We evaluated the test dataset 1000 records at a time for 10 rounds to simulate a real-world scenario where user preferences may change over time. We see that our found that our bandit learner consistently predicts better than static learners – as shown in figure 12. Now if we have access to complete dataset, we can train a better learner which can get lower RMSE values up to 0.93 [4]. However, if we consider real-world implementation, we will only have access to limited data and will need a model that continuously adapts to changing user preferences and helps us get better accuracy of prediction.

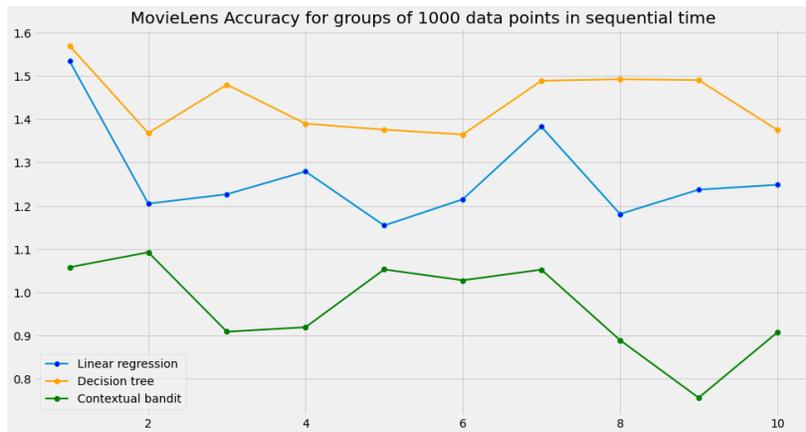
**Figure 11: Comparing static regressors vs incremental learning**

# 8 Conclusion

We reviewed the RL problem and explored contextual bandits as a special abstraction of this. We started with a simple synthetic dataset and showed limitations of static learners to adapt to changing user preferences. We developed a contextual bandits algorithm that builds a SGD learner per arm based on unique context observed. With this approach our learner can continuously adapt to changing preferences and improve prediction accuracy with time. On our synthetic dataset the learner could capture key change points and adjust its weights so that prediction accuracy is improved. The key differentiators here are to have separate learner for each bandit arm and keep adjusting this learner with new values as they come in. Finally, we applied this algorithm to the popular MovieLens dataset for making movie ratings predictions based on user context per genre. We showed a novel approach to use movie genre as the deciding factor for action taken or arm pulled – thus making computation easy rather than considering each movie as the action. Choosing genre as the arm or action or the recommendation and showed 82% accuracy on the MovieLens dataset. Then we compared MovieLens dataset prediction using static learners like linear regression and decision trees. We see that our found that our bandit learner consistently predicts better than static learners. Now if we have access to complete dataset, we can train a better learner which can get lower RMSE values. However, if we consider real-world implementation, we will only have access to limited data and will need a model that continuously adapts to changing user preferences and helps us get better accuracy of prediction.

All our datasets and the contextual bandit library is available open source on our GitHub [12].

# REFERENCES

[1] Li, L., Chu, W., Langford, J., & Schapire, R. (2010). A contextual-bandit approach to personalized news article recommendation. ArXiv, abs/1003.0146

[2] Zhang, C., Wang, H., Yang, S., & Gao, Y. (2019). A Contextual Bandit Approach to Personalized Online Recommendation via Sparse Interactions. PAKDD

[3] Tang, L., Jiang, Y., Li, L., Zeng, C., & Li, T. (2015). Personalized Recommendation via Parameter-Free Contextual Bandits. Proceedings of the 38th International ACM SIGIR Conference on Research and Development in Information Retrieval.

[4] Krishnamurthy, A., Langford, J., Slivkins, A., & Zhang, C. (2019). Contextual Bandits with Continuous Actions: Smoothing, Zooming, and Adapting. COLT.

[5] Majzoubi, M., Zhang, C., Chari, R., Krishnamurthy, A., Langford, J., & Slivkins, A. (2020). Efficient Contextual Bandits with Continuous Actions. ArXiv, abs/2006.06040.

[6] Bietti, A., Agarwal, A., & Langford, J. (2018). A Contextual Bandit Bake-off. arXiv: Machine Learning.

[7] Foster, D.J., Krishnamurthy, A., & Luo, H. (2019). Model selection for contextual bandits. NeurIPS.

[8] Tang, Liang & Jiang, Yexi & Li, Lei & Li, Tao. (2014). Ensemble Contextual Bandits for Personalized Recommendation. RecSys 2014 - Proceedings of the 8th ACM Conference on Recommender Systems. 10.1145/2645710.2645732.

[9] Cañamares, Rocío & Redondo, Marcos & Castells, Pablo. (2019). Multi-armed recommender system bandit ensembles. 432-436. 10.1145/3298689.3346984.

[10] F. Maxwell Harper and Joseph A. Konstan. 2015. The MovieLens Datasets: History and Context. ACM Transactions on Interactive Intelligent Systems (TiiS) 5, 4, Article 19 (December 2015), 19 pages. DOI=http://dx.doi.org/10.1145/2827872, http://grouplens.org/datasets/movielens

[11] Cheng, Weijie & Yin, Guisheng & Dong, Yuxin & Dong, Hongbin & Zhang, Wansong. (2016). Collaborative Filtering Recommendation on Users' Interest Sequences. PLOS ONE. 11. e0155739. 10.1371/journal.pone.0155739.

[12] All the source code for this paper including our normalized datasets and scripts for the bandit learner algorithm, data cleansing and analyzing MovieLens dataset are available at:

https://github.com/dattarajrao/contextualbandits